\documentclass[journal]{IEEEtran}
\ifCLASSINFOpdf
\else
\fi
\newcommand{\ie}{i.e.}
\newcommand{\eg}{\textit{e}.\textit{g}.}
\newcommand\etc{\emph{etc.}}

\usepackage{multirow}
\usepackage{epsfig} 
\usepackage{epstopdf}
\usepackage{graphicx} 
\usepackage{caption}
\usepackage{subfigure}
\usepackage{amsmath,amssymb}  
\usepackage{color}
\usepackage{mathtools}

\begin{document}
%

\title{Leveraging  Structural Context Models and Ranking Score Fusion for Human Interaction Prediction}

\author{Qiuhong~Ke,        
        Mohammed~Bennamoun,
        Senjian~An,
        Ferdous~Sohel,
        and~Farid~Boussaid
\IEEEcompsocitemizethanks{
 
\IEEEcompsocthanksitem Qiuhong Ke,
        Mohammed Bennamoun and Senjian An 
          are with School of Computer Science and Software Engineering,
        The University of Western Australia, Crawley, Australia.\protect\\ 
E-mail:~qiuhong.ke@research.uwa.edu.au, mohammed.bennamoun@uwa.edu.
\protect\\
au, senjian.an@uwa.edu.au
\IEEEcompsocthanksitem Ferdous Sohel is with 
School of Engineering and Information Technology,
Murdoch University, Murdoch, Australia.\protect\\
E-mail: f.sohel@murdoch.edu.au
\IEEEcompsocthanksitem Farid Boussaid is with  School of Electrical, Electronic  and Computer Engineering,
 The University of Western Australia, Crawley, Australia.\protect\\
 E-mail: farid.boussaid@uwa.edu.au
 }}

%
%

\markboth{Submitted to Transaction of Multimedia}%
{Shell \MakeLowercase{\textit{et al.}}: Bare Demo of IEEEtran.cls for IEEE Journals}
%



\maketitle


\begin{abstract}
Predicting an interaction before it is fully executed is very important in   applications such as human-robot interaction  and video surveillance. 
In a two-human interaction scenario, there often  contextual dependency  structure  between  the global interaction  context of the two humans and the local context  of the different body parts of each human. 
In this paper, we propose to learn the structure of the interaction contexts, and combine it with the 
   spatial and temporal information of a video sequence  for a better prediction of the interaction class.
The structural models, including the spatial   and  the temporal  models, are learned with  Long Short Term Memory (LSTM) networks to capture the dependency of 
 the global and local contexts of each RGB frame and each optical flow image, respectively.
  LSTM networks are also capable of  detecting the key information  from the global and local interaction contexts.
Moreover, to effectively combine the    structural models with the  spatial and   temporal      models for interaction prediction, a
 ranking score fusion method is   also introduced to  automatically compute the optimal weight of each model for score fusion.
  Experimental results  on the BIT-Interaction and the UT-Interaction datasets clearly demonstrate the benefits of the proposed method. 
\end{abstract}

\begin{IEEEkeywords}
Interaction Prediction, Interaction Structure, LSTM, Ranking Score Fusion.
\end{IEEEkeywords}

%
\IEEEpeerreviewmaketitle

\section{Introduction}
  \label{intro}

\IEEEPARstart{H}{uman}  interaction prediction, or early recognition,  has a wide range of applications. 
It can help   preventing harmful events (\eg, fighting) in a surveillance scenario. It is also very essential for human-robot interaction (\eg, when a human lifts his/her hand to handshake or opens his/her arms to hug, the robot will then respond accordingly). 
Unlike action recognition whose objective is to classify a full video,    interaction prediction aims to infer an interaction class with a sequence containing only a partial observation of the full interaction activity. Interaction prediction is  challenging 
 due to the large variations in   human postures during a complete  interaction  sequence. 


\begin{figure}[]
\begin{center}
   \includegraphics[width=3.5in]{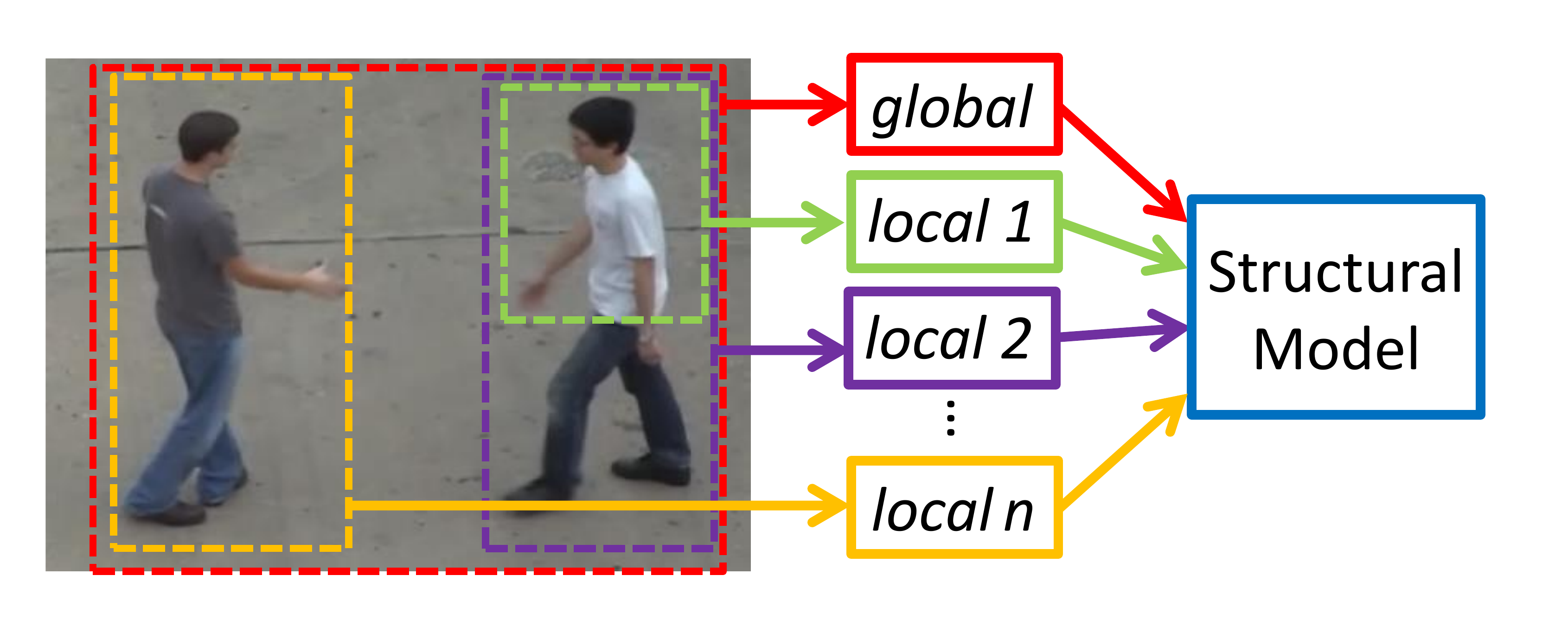}
\end{center}
   \vspace{-3mm}\caption{The structural model 
    aims to capture the contextual dependency and salient information from the  global  and local interaction contexts. 
   }  
\label{fig1}
\end{figure}

Human   interaction involves a sequence of  postures of two humans in a specific scenario. 
The global   interaction
context  of each video frame, containing the two humans,    provides the  overall information of their  positions and relationship. 
 The local context    (including different 
body parts of each human)
provides the fine-grained details of  the body gestures. 

Previous works  focus on using the spatial    and  temporal information   from videos for human interaction prediction \cite{Ryoo2011, lan2014hierarchical, kong2014discriminative, kong2015max, ke2016human}. {Humans usually observe the content of  a scene    from  the global context, to obtain an overall information, followed by an observation of the local    context, to  acquire more details \cite{de2011global}.} We hypothesize that the contextual dependency among the global and local contexts thus
needs to be learned for a better prediction of the interaction class.   {Moreover, the global and local contexts play different roles depending on the types of the interaction.} 
 In the case where 
 the  interaction involves the   movements of both humans (e.g., walking towards   each others or departing), the global context (containing the relationship between the two humans) is more useful  than the  local details. On the other hand,  some other types of  interactions   mainly consist of the movements of a particular local body part of one human (e.g., upper  body part  in ``hugging'' and  lower body part  in ``kicking''). 
 
 To learn the contextual dependency and to capture the salient information   for interaction prediction, we propose to learn structural models from the global and local interaction contexts (as shown in Figure \ref{fig1}).
 More specifically, 
    we  organize the global and local interaction  contexts   in a sequential order and then use       Long Short Term Memory (LSTM)  networks \cite{hochreiter1997long} to process  the sequence and learn the structural models. 
 LSTM networks are designed for  temporal modelling. They  are capable of   detecting salient keywords in sequential data,  such as speech and sentences \cite{sak2014long, palangi2016deep}. 
 LSTM networks have also been used to learn the    spatial dependency and discover salient  local features in images \cite{ varior2016siamese}. We propose to exploit the  structural information of the interaction context by processing the sequence of the global and local contexts  using LSTM networks. This will allow us to capture the   contextual dependency  of the   interaction, and to
 detect the discriminate information, which is relevant to the interaction class   by  ``memorizing'', and to
block the irrelevant information by 	``forgetting''. 
  The learned  structural models   enhance the discriminative power of the global and  local contexts for interaction prediction. Our experiments clearly show the benefits of using structural models for interaction prediction (See Section \ref{expres}).

The temporal information of a video sequence is also very important for  interaction prediction.
Human interactions may last for a long period and can consist of multiple different sub-actions.
It is  insufficient to use a single frame that is  captured before the interaction happens   to infer the   class. 
The temporal  information that is captured along several consecutive  frames, on the other hand, provides   critical cues to  predict a future interaction.
 To extract    the  temporal information, we adopt the temporal convolution network proposed in \cite{ke2016human}, which  learns the temporal evolution using   the  features of several consecutive   optical flow images.  
 We also consider  the   spatial information  of each video frame.  
 
 On this basis,    {our proposed interaction prediction framework} is achieved through the incorporation of   the structural,  temporal  and  spatial   models. 	  {These models have different discriminative abilities in classification.}  
Previous works       average or manually assign weights to fuse the different models \cite{donahue2015long, wang2016temporal}.  To effectively combine the complementary strengths of the proposed structural, spatial and temporal   models, we introduce  a new ranking   score fusion method, which can automatically find the fusion weights of   these models for the final  interaction prediction decision.
  The advantage of the ranking score fusion method over a simple average fusion is shown in Section \ref{expres}.

  
The main contributions of this paper  relate to the proposed learning methodology for interaction prediction. 
\textbf{First}, we   design  novel structural models which are exploited using  LSTM networks to process a sequence of  global and   local interaction contexts   and  improve the performance of    interaction prediction. The structural models learn the contextual dependency and extract the discriminative information that is relevant to the interaction class. 
 {Experimental results clearly demonstrate the benefits of the proposed structural models (Section \ref{expres}).} 
\textbf{Second}, we develop a ranking score fusion method to combine the structural,   spatial,  and   temporal models  for the final prediction of an interaction class. The ranking score fusion method  automatically  finds the optimal weights of these models and is more robust compared to the average fusion approach.
\textbf{Third}, we have   evaluated our method on two interaction datasets and the experimental results demonstrate that the proposed method outperforms  the state-of-the-art methods for human interaction prediction (see Section \ref{expres}).

\section{Related Works}
The proposed method mainly focuses on  learning sequences of global and local contexts   for human  interaction prediction. 
 In this section, we therefore briefly describe existing works on action prediction and sequence learning.
 
Ryoo \cite{Ryoo2011} presented one of the early works on human interaction prediction. This work formulated an interaction prediction process as a posterior probability and  represented the video frames with integral  bag-of-words (IBoW) and dynamic bag-of-words (DBoW) to model the temporal evolution of features. Hoai and De la Torre \cite{hoai2014max} proposed   a structured output SVM to train a detector to recognize partial events. When testing on action data, they used the Euclidean distance transform of binary masks between frames to create a codebook and computed  a histogram of temporal words to represent  a sequence of frames. Cao et al. \cite{cao2013recognize} divided each activity into multiple ordered temporal segments and constructed  a matrix basis for each segment with the  spatio-temporal features from the training data. A sparse coding method (SC) is then used to  approximate the features of the test video with one matrix basis   or a mixture of several matrix bases (MSSC). Lan et al. \cite{lan2014hierarchical} introduced a ``Hierarchical Movemes'' (HM) feature (\ie, combining features  from coarse to fine temporal levels based on HOG, HOF and MBH features) as representations and used SVM to jointly learn the appearance models at different levels and their intra relationships to predict interaction.  Kong et al. \cite{kong2014discriminative} represented  partial videos using  bag-of-words features, and learned the multiple temporal scale support vector machine (MTSSVM) based on a structured SVM to recognize unfinished videos.  Kong et al. \cite{kong2015max} extended their work and proposed a max-margin action prediction machine (MMAPM) for early recognition of unfinished actions. The limitation of the above mentioned  methods lies in their reliance on  low-level features. Recently, Ke et al. \cite{ke2016human} applied CNNs on flow coding images to learn the temporal information
for human interaction prediction. This method uses only temporal features and lacks the discriminative spatial features of human postures.

Sequence learning has been used in  the temporal domain to capture the  temporal information associated to  consecutive frames  in a video. Traditional sequential models such as Hidden Markov models (HMMs) \cite{eddy1996hidden} and Conditional Random Fields (CRFs)\cite{lafferty2001conditional} have been successfully  used for action recognition \cite{ chung2008daily, zhang2010action, tang2012learning, song2013action}.  However, they are not suitable for applications with high dimensional features \cite{Ryoo2011}. Besides, they are not designed to learn long-term dependencies. 
LSTM networks \cite{hochreiter1997long}, on the other hand, are capable  to learn long-term dependencies. 
LSTM networks  \cite{hochreiter1997long} is a  variant of Recurrent Neural Networks (RNNs) \cite{jaeger2002tutorial}  with LSTM cells, which   can remove or add information   over a period of time.    LSTM ntworks have been successfully applied for   speech recognition \cite{graves2013speech},  video description and recognition  \cite{ma2016learning, donahue2015long, liu2016spatio, liujun2015cvpr}.  
 LSTM networks have  also been   used in   salient  keywords detection in sentences for document retrieval \cite{palangi2016deep}. Although RNNs and LSTM are designed for temporal modelling, they have also been used 
to    process   sequences of local features in images to exploit the spatial dependency and extract discriminative information  for powerful image representation \cite{shuai2016dag, zuo2015convolutional, varior2016siamese}. 

{\color{black}{  



\section{Proposed Approach}


 \begin{figure*}[h]
\vspace{0mm}
\begin{center}
   \includegraphics[width=6.5in]{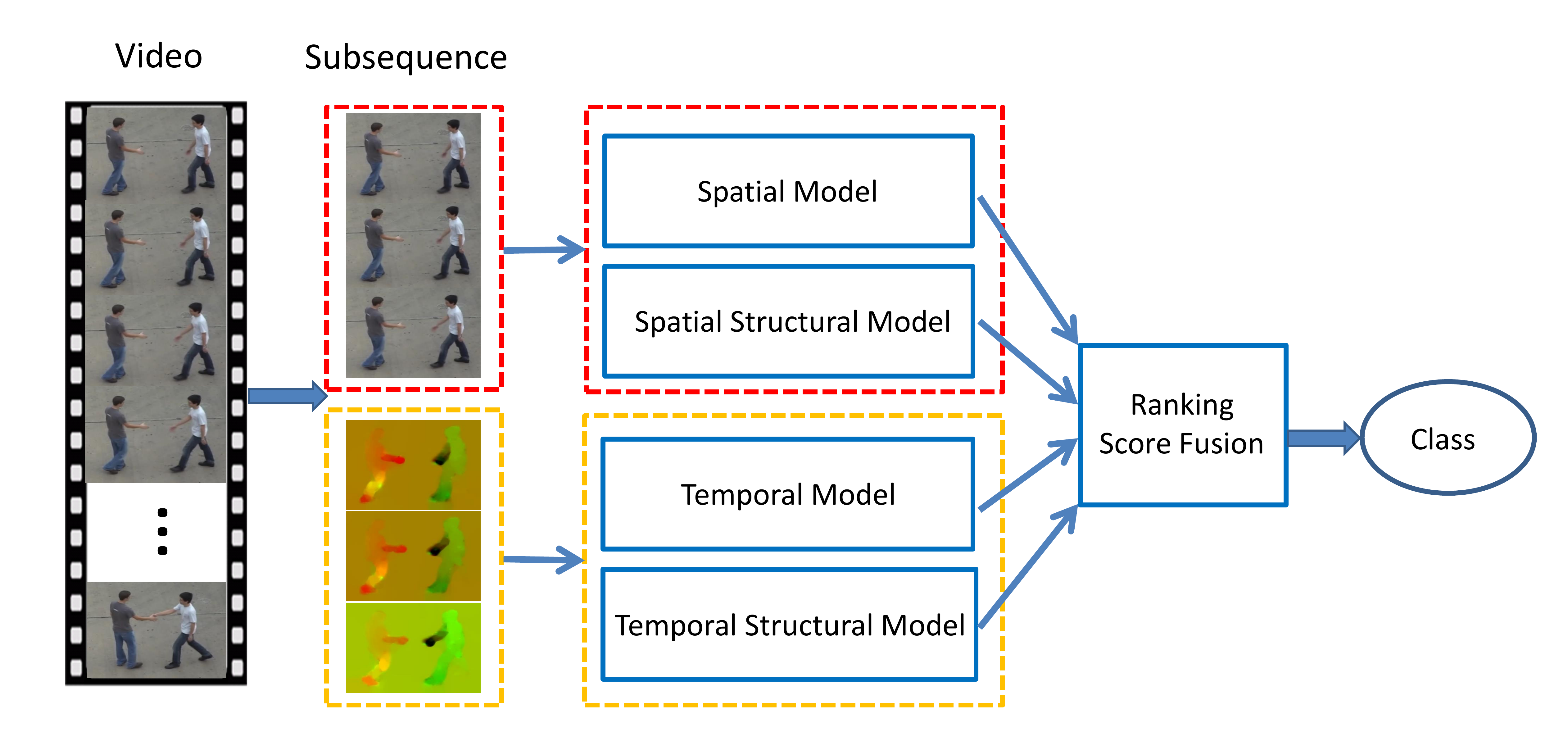}
\end{center}
      \vspace{-5mm}\caption{Outline of the proposed method.  The goal of the proposed method is to effectively combine the   spatial, spatial structural, temporal and temporal structural
      information to predict the interaction class from a subsequence  
       containing only a partial observation
      of the interaction.   We first compute optical flow images from consecutive video frames  of the partial sequence.   The video frames are fed to the spatial    and the   spatial structural models, while the optical flow images are fed to the temporal    and the   temporal structural models.
       The output   scores of these models are fused with a ranking score fusion method to predict the   class of the partial observation of the interaction.}
   \label{overall1}
\vspace{0mm}
\end{figure*} 

 An overall architecture of the proposed method is shown in 
  Figure~\ref{overall1}. 
It contains a spatial model, a temporal model and 
two   structural models (i.e., a spatial structural model   and a temporal structural model).  The prediction scores of the four models are fused using  a ranking score fusion method for the final 
decision of the interaction class.
The goal of the proposed method is to effectively combine the structural, spatial  and temporal   information of the videos for interaction prediction.
Learning the spatial and temporal information is a common practice for video action recognition. 
For the  scenario of two-human interaction, we introduce structural models to extract the salient information related to the interaction class and  to learn the 
 contextual dependency among the global context of the two humans and their local body gestures.
In this section, we first describe the proposed structural models in details, and then briefly introduce the  spatial and temporal  models. Finally, we   present the  ranking score fusion   and the testing methods.

\subsection{Structural Models}

\begin{figure*}[h]
\vspace{0mm}
\begin{center}
   \includegraphics[width=6.8in]{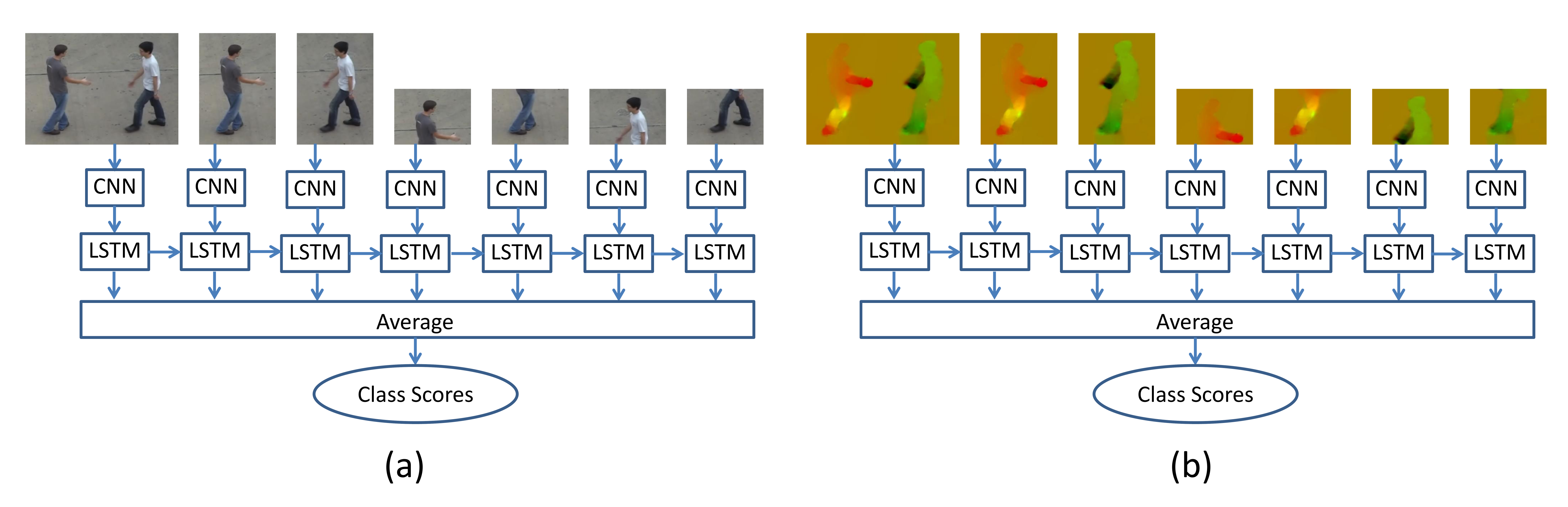}
\end{center}
      \vspace{-2mm}\caption{The Network architecture of the structural models: (a) Spatial Structural Model and (b) Temporal Structural Model.}
   \label{lstm1}
\vspace{0mm}
\end{figure*}

In a two-human-interaction context, the global   
context    provides the overall information of the  positions and gesture relevance of the two humans.
For example, if the two humans  walk towards   each other to   handshake (as shown in Figure \ref{fig1}), their  movements of body gestures     are   similar and  {the distance of the two actors  becomes smaller}. While   if someone intends to kick another person,    the gestures of the two humans  are generally different.
  The local context   including  each human and  their     upper  and   lower body parts
  provides fine-grained details of the gestures. There exist intrinsic relationships and dependencies among    the global and local contexts. In addition,  the global and local contexts play different roles for different interactions.  In some interactions,  {both humans perform actions}. The global context with regard to the relationship of the two humans provides discriminant information  to distinguish between the different types of interactions (e.g., walking towards or departing from each other). For some other interactions,   {only one actor performs an action}, while the other actor stands still. In this case, the local context of the moving human provides more details than the global context and is thus more important. The upper and lower body parts     also a have different importance  depending on the types of    interactions (e.g., the upper body part is more important in ``hugging'' while the lower part is more important in ``kicking'' action). 
 Considering that LSTM networks are capable to learn contextual dependencies and detect salient information in sequences  \cite{palangi2016deep,  shuai2016dag, zuo2015convolutional, varior2016siamese}, we propose,  in this paper, to organize the global and local interaction contexts  in a sequence and use LSTM networks to  learn the sequence to derive the structural models for a good understanding of an   interaction class.

 \subsubsection{Model Input}
 \label{networkin}
  As shown in Figure~\ref{lstm1}, the proposed structural models include the spatial structural     and the temporal structural models. These models aim to learn the contextual dependency and detect salient information from 
  the global and local contexts of the frame and optical flow images, respectively.  
  The global context is   the image region which contains both humans, while the local context consists of  6 local patches, including the whole body, the upper and the lower  body parts of each human. 
  Inspired by the theory that the human visual system analyses the contents of visual scenes sequentially from  the global  to the local  scene contexts \cite{de2011global}, we organize the global and local contexts in a sequential order,  as shown in Figure~\ref{lstm1}. More specifically, the order is:  the global context with two humans, the whole body of the left human, the whole body of the right human, the upper body part of the left human, the lower body part of the left human, the upper body part of the right human, the lower body part of the right human.
  
  For the temporal structural model,  the inputs are the global and local contexts of the optical flow image, which is derived from  the optical flow between  consecutive frames.
   The $x$ and the $y$ components of an optical flow vector are scaled  to values between  0 and 255 using a linear transformation. The two components ($x$ and $y$) correspond to  two channels of the optical flow image, and the third channel of the optical flow image is set to 0.

\subsubsection{CNN Feature Representation}
\label{rep}
The global and local contexts of the frame and optical flow images are fed to   CNN models that are pre-trained on ImageNet \cite{deng2009imagenet} to extract   fixed-dimension   features.  
  CNN pre-trained models  have been shown to be transferable across domains \cite{razavian2014cnn}, and have achieved a better performance than hand-crafted features    in a variety of visual recognition tasks  \cite{li2015feature, liu2015crf, ke2014rotation, ke2017new,  ke2017skeletonnet}. Particularly,  CNN pre-trained models  have been successfully used to 
extract features from video frames and optical flow images for action recognition and detection  \cite{simonyan2014two, yue2015beyond, gkioxari2015finding}. 
In this paper, the CNN-M-2048 model \cite{chatfield2014return} is used for feature extraction due to its successful application in action recognition \cite{simonyan2014two}.  The convolutional layers of this CNN model contain  96 to 512 kernels with a size varying from $3\times3$ to $7\times7$. The rectification unit (ReLU) \cite{nair2010rectified} is used as a nonlinear activation function. 
The output of the first fully connected layer (layer 19)  of the network is used as the feature representation of the  input image.

\subsubsection{Model Learning}
  The CNN features of the global and local contexts are fed to LSTM networks in a sequential order as described in Section \ref{networkin}. 
LSTM  netwoks \cite{hochreiter1997long} are RNNs \cite{jaeger2002tutorial}  with LSTM cells.  Standard RNN can be regarded as multiple copies of  the same network, allowing for processing time series. The traditional RNNs suffer from the problems of  vanishing    and exploding gradients \cite{hochreiter2001gradient, pascanu2012difficulty}. 
Compared to RNNs, LSTM networks contain  memory cells,  which are capable to learn long-term dependencies. The LSTM cell is composed of a forget gate $f_t$, an input gate $i_t$, a cell state $C_t$, an output gate $o_t$ and a hidden state  $h_{t}$. 
Given an input  $x_t$ at time step $t$ and a hidden value $h_{t-1}$ of the previous time step $t-1$, the equations of each gate and the states follow the equations below.

\begin{equation}
\label{eqlstm}
\begin{array}{c}
f_t=\sigma(W_fx_t+U_fh_{t-1}+b_f) \\
\widetilde{C_t}=\tanh\left(W_cx_t+U_ch_{t-1}+b_c\right) \\
i_t=\sigma(W_ix_t+U_ih_{t-1}+b_i)\\
C_t=i_t*\widetilde{C_t}+f_t*C_{t-1} \\
o_t=\sigma(W_ox_t+U_oh_{t-1}+V_oC_t+b_o) \\
h_t=o_t*tanh(C_t) \\
\end{array}
\end{equation}

\noindent where $W_f$, $U_f$, $W_c$, $U_c$, $W_i$, $U_i$, $W_o$, $U_o$ and $V_o$
are the weight matrices and $b_f$,  $b_c$, $b_i$ and $b_o$ are the biases. $\sigma(\cdot)$ is the sigmoid function. 

 At each time step $t$, the LSTM updates the hidden value and output value with the above equations. By determining when to remember and when to forget, LSTM networks are capable to learn dependencies and to detect the discriminate information in  a sequence. }

 As shown in Figure \ref{lstm1},  each sequence  feeding to LSTM networks contains seven time steps of CNN features. LSTM networks output a hidden value and an output value at each time step using Equation (\ref{eqlstm}).
  The outputs of all time steps are fed to a hidden  layer including a fully connected (FC) layer and ReLU, followed by another FC layer and a softmax layer to generate class scores for each steps. 
  More specifically, let the  output value of LSTM cell at time step $t$ be $o_t \in \mathbb{R}^d$. $d$ denotes the number of hidden units of the LSTM cell. The probability score of the $i^{th}$ class at the step $t$ is given 
  by:
  
  \begin{equation}
  \begin{array}{c}
   z_t = W_2(\max(W_1o_t+b_1)+b_2\\
   \\
 y_{t,i}= \dfrac{\exp z_{t,i}}{\sum\limits_{j=1}^m \exp z_{t,j}} 
  \end{array}  
  \end{equation}
  
 \noindent where $z_t \in \mathbb{R}^m$ is the   vector   fed to the softmax layer. $W_1 \in \mathbb{R}^{d_1\times d}$ and $b_1 \in \mathbb{R}^{d_1}$ denote  the parameters of the first FC layer. $d_1$ is the number of units of the first FC layer.
  $W_2 \in \mathbb{R}^{m\times {d_1}}$ and $b_2 \in \mathbb{R}^m$ denote  the parameters of the second FC layer.
  $m$ is the number of interaction  classes. The class scores of all time steps are averaged to produce the final probability  scores of the structural model.

\subsection{ Spatial and Temporal  Models}

\begin{figure}[h]
\vspace{0mm}
\begin{center}
   \includegraphics[width=3.5in]{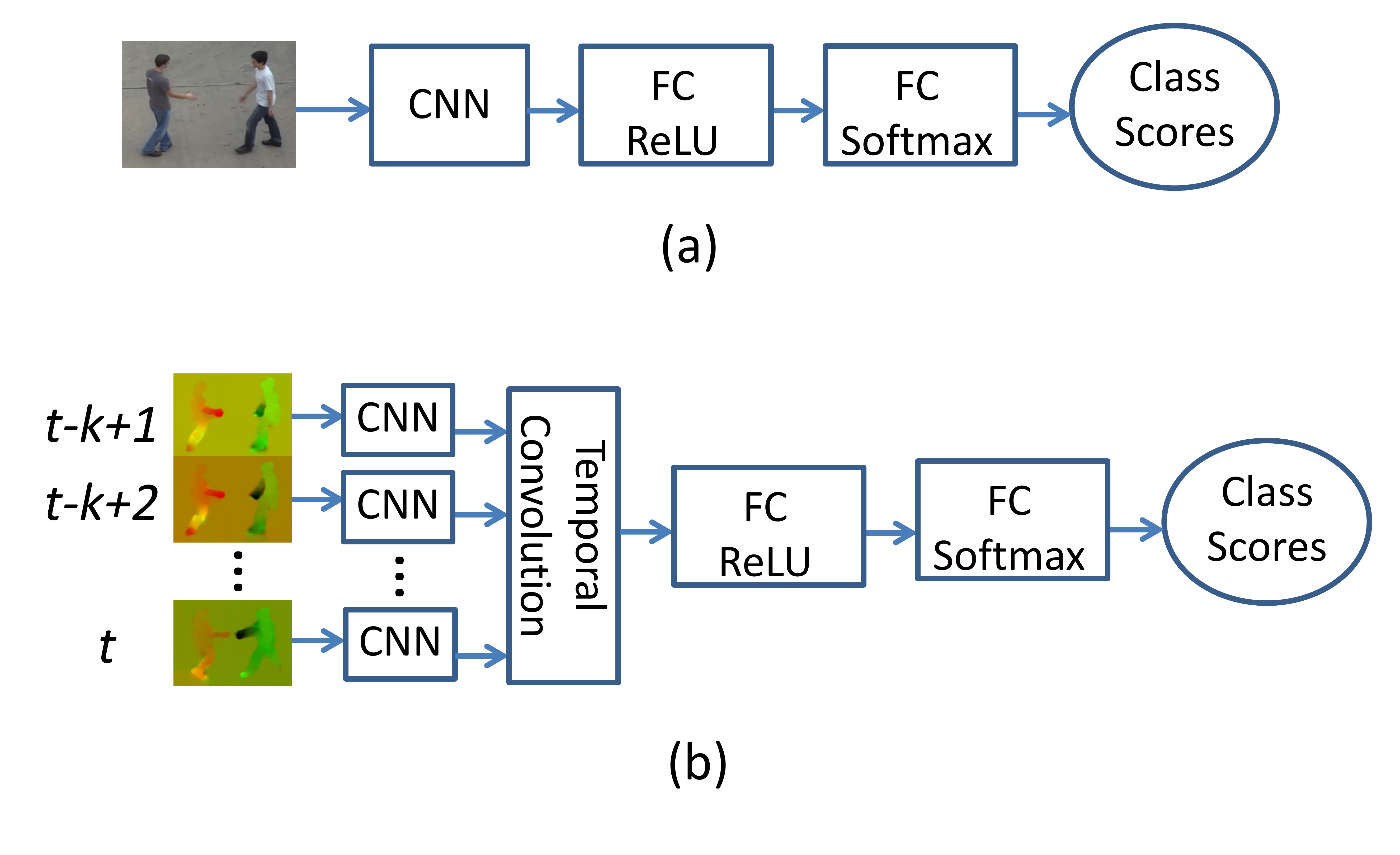}
\end{center}
      \vspace{-2mm}\caption{ Network architecture of  (a)  the spatial   model and (b) the  temporal   model.
      ``FC'' denotes a ``fully connected'' layer.
      During the training of the temporal model, every stack of $k$   optical flow images is fed to the model. During testing,
      the prediction scores of the $t$ time step is generated by feeding the sequence formed by the $t^{th}$   and the previous $k-1$ optical flow images to the model. The current $t^{th}$ optical flow image is repeated when $t<k$.}
   \label{st}
\vspace{0mm}
\end{figure}


 The     spatial   interaction  context   provides the static human postures of the two interacting humans.  
 The  spatial model   is introduced  to extract the spatial information of the interaction context  from each individual frame for interaction prediction. As shown in Figure~\ref{st}(a), the input of the spatial model  is a single frame image 
   containing both humans.  The   CNN-M-2048 model \cite{chatfield2014return} is used to extract a feature from the input frame image which is the same as the feature representation of the structural models (see Section \ref{rep}).  The feature is then fed to a      fully connected (FC)  layer,  a ReLU,    another FC layer  and  a softmax layer to generate the probability scores of an interaction.

Because human interaction involves the movements of human limbs along a sequence, using only the spatial feature of each individual frame is insufficient to infer  {the interaction class}, while the temporal feature of multiple consecutive frames provides more information. The goal of the temporal model is to exploit the temporal information for interaction prediction. 
For the interaction prediction task, which aims to recognize an interaction class at an early temporal stage before an 
interaction happens, the   testing sequences consist of subsequences containing only a  partial observation of the full activity.
The temporal evolution of partial sequences needs to be learned for interaction prediction. The temporal
convolution network \cite{ke2016human} aims to model the temporal information from subsequences and has been successfully 
used for interaction prediction.
We therefore adopt the temporal
convolution network \cite{ke2016human} to build our temporal model. 
 As shown in Figure~\ref{st}(b), the input of the temporal model is  a set of consecutive  optical flow images, which are fed to a pre-trained CNN model to  extract  CNN features similar to the spatial and structural models. The sequence of   features are then processed with a temporal convolution layer.    The  output feature is   a compact feature vector of the sequence. It is then fed to a hidden layer and  an output layer consisting of a FC and a softmax layer  to produce the probability scores of an interaction class.   During testing, the prediction result of the time step $t$ is derived by feeding a sequence consisting of the current $t^{th}$    and the previous $k-1$     optical flow images to the model ($k$ denotes the number of    consecutive optical flow images used for training). The current flow image is repeated for $k-t$ times to generate a sequence of length $k$ when $t<k$.

\subsection{Ranking Score Fusion}

For different datasets, the relative importance of the spatial structural, temporal structural, spatial   and temporal models may vary due to their different  discriminations between classes. 
  The proposed ranking score fusion method is used to find the optimal fusion weights to combine these models. Given a testing 
sequence, the four models generate a  score matrix  $S \in \mathbb{R}^{4\times m}$ at each time step ($m$ is the number of interaction classes):
 
\begin{equation}
S\triangleq\left[\begin{array}{cccc}
s_{11},~s_{12},~\cdots,~&s_{1m}\\
s_{21},~s_{22},~\cdots,~& s_{2m}\\
s_{31},~s_{32},~\cdots,~& s_{3m}\\
s_{41},~s_{42},~\cdots,~& s_{4m}\\
\end{array}\right]  
\label{equscore}
\end{equation}
\noindent where   $s_{pq}$ denotes the score of the $p^{th}$ model for the $q^{th}$ interaction class.  Each column of $S$ corresponds to one class.  Inspired by the ranking theory \cite{joachims2002optimizing}, we learn to rank these columns  so that $\mathbf{s}_{l} \succ \mathbf{s}_j$, where  $l$ is the ground truth class label of the video, $\mathbf{s}_{j}$ is the $j^{th}$ column of $S$, and $\succ$ denotes the order between two vectors. 
The task is to learn a linear  function $g: \mathbb{R}^4 \mapsto \mathbb{R}$ which induces an ordering on the columns, i.e.,
\begin{equation}
 g(\mathbf{s}_{l}) \succ g(\mathbf{s}_{j}),  j =1,\cdots,m~~and~~j\neq l 
\end{equation}
where
\begin{equation}
g(\mathbf{\mathbf{s}}) = \mathbf{w}^T \mathbf{s}
\end{equation}
and $\mathbf{w} \in \mathbb{R}^{4}$ is a four dimensional weight vector of the linear function.  

Let $\{\mathbf{x}^{(1)},\mathbf{x}^{(2)}\}$ be a pair of  columns of a score matrix, we have either $g(\mathbf{x}^{(1)}) \succ g(\mathbf{x}^{(2)})$ or $g(\mathbf{x}^{(2)}) \succ g(\mathbf{x}^{(1)})$, or equivalently 
\begin{equation}
 y \mathbf{w}^T (\mathbf{x}^{(1)} - \mathbf{x}^{(2)}) > 0
\end{equation}
\noindent where
\begin{equation}
y\triangleq \left \{
\begin{array}{ll}
+1,  \quad if \quad \mathbf{x}^{(1)} \succ \mathbf{x}^{(2)}\\
-1,  \quad if \quad \mathbf{x}^{(2)} \succ \mathbf{x}^{(1)}.
\end{array} \right.
\end{equation}

\noindent Thus one can treat the difference vector  $\mathbf{x}^{(1)}-\mathbf{x}^{(2)}$  as a training example with label $y$. The weight vector $\mathbf{w}$ can be obtained by training a binary classifier. More precisely, by using the score matrices of all the videos at every time step, we can create a training set
\begin{equation}  
Q = \left\{ \left( \mathbf{x}_i^{(1)} - \mathbf{x}_i^{(2)} \right), y_i  \right\}_{i=1}^N
\end{equation}
 where   $N$ is the total number of the training pairs generated from the score matrices.

The weight vector  $\mathbf{w}$ can be obtained by solving the following optimization problem:
\begin{equation}
\begin{aligned}
& \underset{\mathbf{w}}{\text{argmin}}
& & \Arrowvert \mathbf{w} \Arrowvert ^2 + C \sum_{i=1}^n \epsilon_i \\
& \text{subject to}
& & y_i \mathbf{w}^T (\mathbf{x}_i^{(1)} - \mathbf{x}_i^{(2)} ) \geqslant 1- \epsilon_i, i=1,\cdots,N. \\
\end{aligned}
\end{equation}

However, this may result in negative weights (which  is  counter-intuitive for a model to contribute negatively). To ensure that the contribution of each model is   zero at worst, a non-negative constraint on the weight vector $\mathbf{w}$ is added. The problem is solved approximately using the following iterative method: first,   a weight vector without the  constraint is obtained, then the negative weights are set  to   zero and  the remaining weights are re-trained until all of the weights become non-negative.

\subsection{Sequential-level Prediction}

Given the trained four classification models and  a testing sequence containing $n$ frames,   there will be a score matrix $S \in \mathbb{R}^{4\times m}$ as described in Equation (\ref{equscore}) at each time step of the sequence. 
Let the weights learned by the ranking score fusion method be $\mathbf{w}_o \in \mathbb{R}^4$.
 The four rows of $S$ are then combined using   $\mathbf{w}_o$ to produce the final scores of each time step. 
 Let the final score of    time step $t$  (after combing the four models) be $\mathbf{c}_t=[c_{t,1},c_{t,2},\cdots, c_{t,m}]$.  $c_{t,i}=\mathbf{w}_o^T\mathbf{s}_i$ is the final score for class $i$. $\mathbf{s}_i\in \mathbb{R}^4$ is the  $i^{th}$ column of 
 $S$, which consists of the scores of the four models for class $i$. $m$ corresponds to the number of classes. 
  The prediction label of  the sequence at time step $t$ is then identified as class $p_t$  where
\begin{equation}
p_t=\arg\max_{i} c_{t,i}
\end{equation}   
Now let $\mathbf{h}=[h_1,h_2,\cdots, h_m]$ be the histogram of the set $\Omega\triangleq \{p_t: t=1,2,\cdots, n\}$, where $n$ is the number of frames of the sequence and $h_i$ is the number of elements of $\Omega$ that is equal to $i$. Finally, the sequence is predicted as class $p^{*}$ where
\begin{equation}
p^*=\arg\max_i h_i
\end{equation} 
This method is called majority vote,  which determines the class label of a sequence by counting the predicted labels at all time steps.

\section{Experiments}
\label{comp}
 
In this section, we present our evaluation results on two datasets that have been used for  interaction prediction.

\subsection{Datasets}

The proposed method has been evaluated on two interaction datasets, \ie,
the BIT-Interaction dataset \cite{kong2012learning}  and the UT-Interaction dataset \cite{Ryoo2011}. 

\textit{BIT-Interaction dataset} \cite{kong2012learning}: This dataset   consists  of 8 types of interactions between two humans (bow, boxing, handshake, high-five, hug, kick, pat and push). Each class of interactions contains  50 videos.  It is a very challenging dataset, including variations in illumination conditions, scales, subject appearances and viewpoints. In addition, there  are also    occlusions   by  holes, bridges, pedestrians, \etc.

\textit{UT-Interaction dataset} \cite{Ryoo2011}: This  dataset  includes two sets. The background of Set 1 is simpler and mostly static. In contrast, the background is complex and slightly moving on Set 2. Each set includes 60 sequences of  videos belonging to 6 interaction classes, \ie, handshake, hug, pointing, kick, push and punch. 

For each testing video, the interaction is predicted  in 10 observation ratios, from 0 to 1, with a step size of  0.1.
In other words, each  testing video is divided into 10 partial sequences. The $i^{th}$ sub-sequence   consists of the frames of  $[1: round(\frac{ni}{10})]$, where $n$ is the number of frames in the full video. 
A prediction accuracy under an observation ratio of 0.2 denotes  the accuracy that is tested with the the second sub-sequence containing frames   $[1:round(0.2n)]$. If the observation ratio is  1, the accuracy is tested with the entire video.
 During training,  the two humans in each frame are detected using the detector in \cite{ess2008mobile}. The image region with both human regions is used as the global region to train the models.
 In both datasets, the numbers of units of the LSTM and the hidden layers in the structural models are set to 512 and 128, respectively.   To train the temporal model, every seven consecutive optical flow images are used as   input. The  number of units of the hidden layer in the  spatial and temporal models is set to 512.     

\subsection{Experimental Results}
\label{expres}

The proposed method is compared with the previous methods.  For both datasets,  the same testing protocol that was used in the   previous works was adopted for a fair performance comparison. 
 In addition, the following baseline are also conducted 
to show the benefits of the structural models and the ranking score fusion method.


\textit{Spatial+Temporal+Ranking} (\textbf{Sp\_Tp\_Rank}):  the prediction scores of the spatial model and the temporal model are fused using the weights learned by the ranking score fusion method to produce the final prediction. Compared to the proposed method, this baseline does not include the proposed structural models.
   
\textit{Spatial+Temporal+Structural+Average} (\textbf{Sp\_Tp\_St\_Avg}):   the prediction scores of the spatial, the temporal and the proposed structural models are averaged to generate the prediction results. Compared to the proposed method, this baseline does not include the ranking score fusion method.


\subsubsection{Result on the BIT-Interaction Dataset}
There are   400   videos in this dataset. Following the same procedure  as in   \cite{kong2012learning}, a random sample of 272 videos are used for training and the remaining   videos are used for testing.

 \begin{figure}[h]
 \vspace{0mm}
\begin{center}
   \includegraphics[width=3.5in]{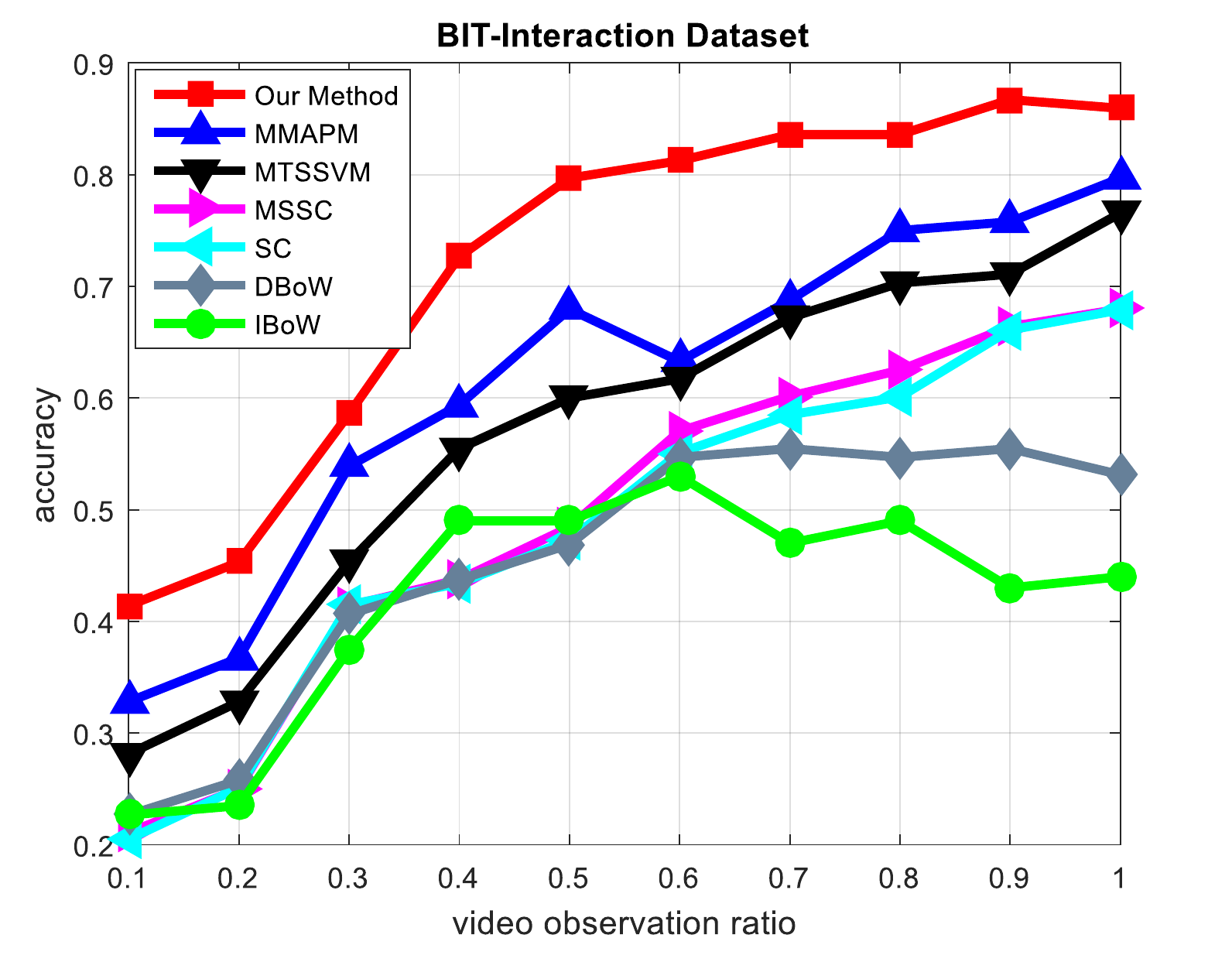}
\end{center}
  \caption{Performance comparisons of the proposed method with other   methods on  the BIT-Interaction dataset. 
  (Best viewed in color)} 
\label{bitcmpall}  
  \end{figure}

    The proposed method is compared with other methods, including IBoW \cite{Ryoo2011} DBoW \cite{Ryoo2011}, SC  \cite{cao2013recognize}, MSSC \cite{cao2013recognize}, MTSSVM \cite{kong2014discriminative} and   MMAPM \cite{kong2015max}. The results are  shown in Figure \ref{bitcmpall}. It can be seen that  the proposed method performs consistently better than other methods for all observation ratios.  When using 20\% of the video (\ie, observation ratio is 0.2) to predict the interaction class, the performance of the proposed method is 45.31\%, which is 8.59\% better than MMAPM \cite{kong2015max} (36.72\%). 
    When   testing with half sequences (\ie, observation ratio is 0.5), the proposed method achieves an accuracy of 79.69\%. Compared to MMAPM \cite{kong2015max}(67.97\%), the improvement is 11.72\%.

   Table \ref{tabbit} shows the comparisons of the proposed method with baseline Sp\_Tp\_Rank. It can be seen that
   the proposed method achieves a better performance in 9 out of 10 cases. When using only 10\%  of each testing 
   video to predict the interaction class, the performance of the proposed method is 41.41. Compared to  Sp\_Tp\_Rank(39.06),
   the improvement is 2.35\%. When testing with half sequences (\ie, observation ratio is 0.5), the improvement of the 
   proposed method  is  3.13\% (from 76.56\% to 79.69\%). Compared to Sp\_Tp\_Rank， the proposed method incorporates
   the proposed structural models. It clearly shows the benefits of the proposed structural models. The structural models  learn the contextual relationships between the global and local contexts, and   
   detect the  discriminate salient features that are related to the interaction class. This provides  complementary information  
   to the spatial and temporal models and improves the interaction accuracy. 
      
    The comparison of 
   the proposed method with baseline  Sp\_Tp\_St\_Avg is also shown in Table \ref{tabbit}. It can be seen that the  proposed method performs better than  Sp\_Tp\_St\_Avg  in all observation ratios. When testing with an observation ratio of 0.1, the accuracy achieved by the proposed method is 7.82\% better than Sp\_Tp\_St\_Avg (\ie, from 33.59\% to 41.41\%). The improvement is more significant with observation ratio 0.5, with an improvement of  10.16\% (from 69.53\% to 79.69\%). Both the proposed method and Sp\_Tp\_St\_Avg  incorporate the same models. Sp\_Tp\_St\_Avg   combines the spatial, structural and temporal models with the average fusion method. These models are learned with different features, which have 
   different discriminations between classes. Simply averaging these models thus generate suboptimal results. The proposed
   method uses the proposed ranking method to find the optimal fusion weights between these models  and produces better results. These significant improvements of the proposed method clearly show the advantage of the ranking score fusion method.

   Table \ref{tabbit2} compares the performance of the proposed method with the four individual models. It can be seen 
   that with this dataset, the temporal and temporal structural models perform better than the spatial model and  the spatial structural model. When the observation ratio is 0.5, the prediction accuracy of the temporal and temporal structural models are   more than 40\% better than the spatial and spatial structural models.  The temporal and temporal structural models are learned with the optical flow images, while the spatial
   and spatial structural models are learned with the video frames. The backgrounds of the video frames of this dataset are very complex.  {The optical flow images  are generated from the motion of two consecutive frames. The 
    background noise is removed, resulting a better  prediction accuracy.} The weights learned by the score ranking method is $[0, 0.47, 0,   0.53]$ for the spatial, temporal, spatial structural and temporal structural models. It can be seen that the fusion of these four models using the learned weight improves the performance of each model with most observation ratios.

{\renewcommand{\arraystretch}{1.5}
\setlength{\tabcolsep}{8pt}
\begin{table*}[h]
\begin{center}
\caption{Performance comparison of the proposed method with  other baselines on BIT-Interaction dataset. } 
\label{tabbit}
\begin{tabular}{ccccccccccc}
\hline
 
\multirow{2}{*}{Methods} &
      \multicolumn{10}{c}{Observation Ratio} \\
        &0.1&0.2&0.3 & 0.4 & 0.5 & 0.6 & 0.7 & 0.8 & 0.9 &1.0 \\
\hline

        Sp\_Tp\_Rank&39.06\%&42.19\%&57.81\%&70.31\%&76.56\%&79.69\%&81.25\%&\textbf{85.16\%}&\textbf{86.72\%}&84.38\%\\
    Sp\_Tp\_St\_Avg&33.59\%&39.84\%&51.56\%&60.16\%&69.53\%&72.66\%&73.44\%&73.44\%&75.00\%&73.44\%\\
  Proposed Method &\textbf{41.41\%}&\textbf{45.31\%}&\textbf{58.59\%}&\textbf{72.66\%}&\textbf{79.69\%}&\textbf{81.25\%}&\textbf{83.59\%}&83.59\%&\textbf{86.72\%}&\textbf{85.94\%}\\

\hline
\end{tabular}
\end{center}
\end{table*}
}

 {\renewcommand{\arraystretch}{1.5}
\setlength{\tabcolsep}{7.5pt}
\begin{table*}[h]
\begin{center}
\caption{Performance comparison of the proposed method with individual models on BIT-Interaction dataset. } 
\label{tabbit2}
\begin{tabular}{ccccccccccc}
\hline
 
\multirow{2}{*}{Methods} &
      \multicolumn{10}{c}{Observation Ratio} \\
        &0.1&0.2&0.3 & 0.4 & 0.5 & 0.6 & 0.7 & 0.8 & 0.9 &1.0 \\
\hline
Spatial&15.62\%&18.75\%&21.88\%&30.47\%&31.25\%&35.16\%&33.59\%&34.38\%&33.59\%&33.59\%\\
   Temporal \cite{ke2016human} &39.06\%&42.19\%&57.81\%&70.31\%&76.56\%&79.69\%&81.25\%&\textbf{85.16\%}&\textbf{86.72\%}&84.38\%\\
    Spatial Structural &21.88\%&23.44\%&25.00\%&28.12\%&30.47\%&32.03\%&32.81\%&33.59\%&32.81\%&32.03\%\\
   Temporal Structural& 35.16\%&44.53\%&56.25\%&67.97\%&\textbf{79.69\%}&79.69\%&80.47\%&82.03\%&82.81\%&82.03\%\\
   
  Proposed Method &\textbf{41.41\%}&\textbf{45.31\%}&\textbf{58.59\%}&\textbf{72.66\%}&\textbf{79.69\%}&\textbf{81.25\%}&\textbf{83.59\%}&83.59\%&\textbf{86.72\%}&\textbf{85.94\%}\\

\hline
\end{tabular}
\end{center}
\end{table*}
}

 \begin{figure*}
 \vspace{0mm}
\begin{center}
   \includegraphics[width=6.5in]{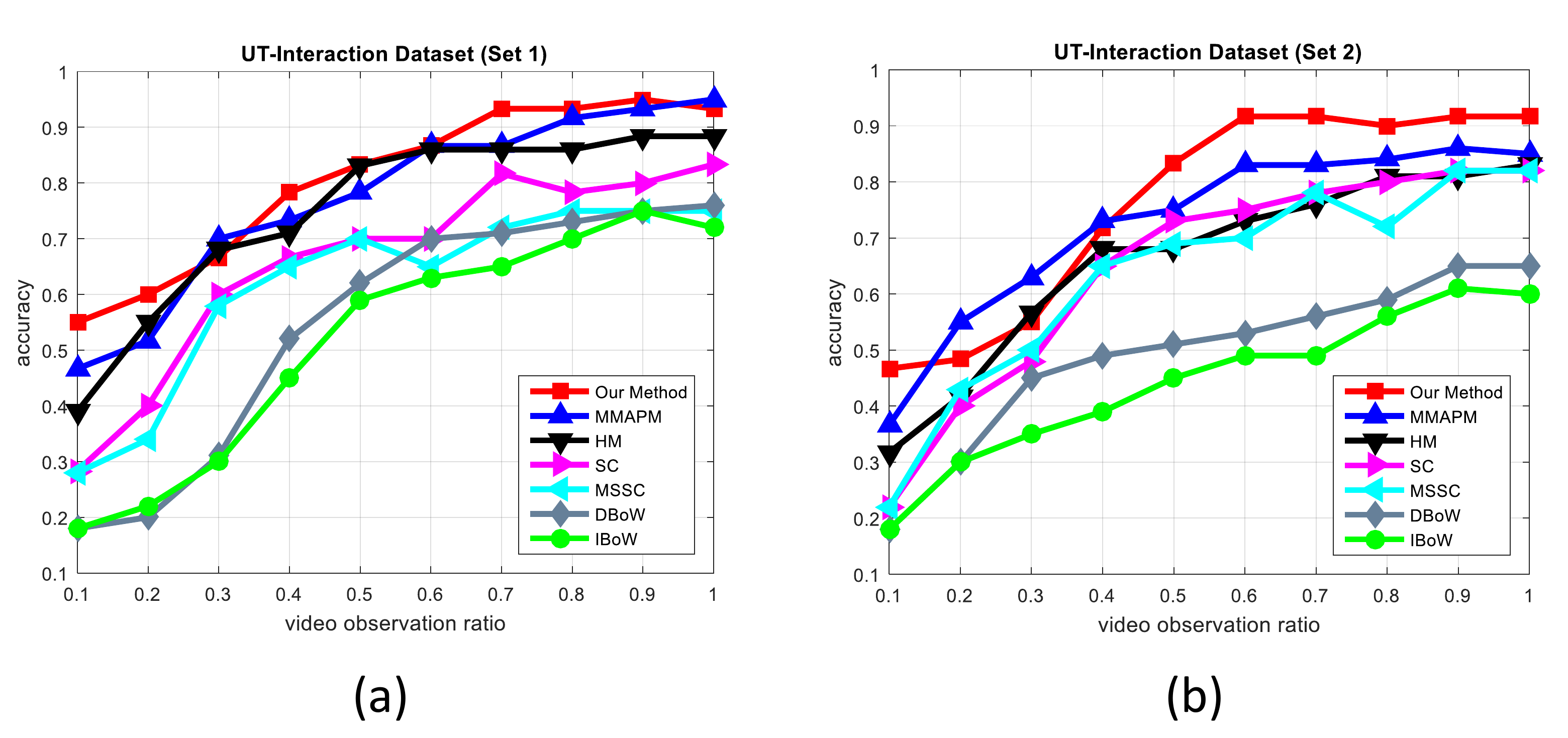}
\end{center}
  \vspace{0mm}\caption{
Performance comparisons of the proposed method with other   methods on  the UT-Interaction dataset (a) Set 1 and (b) Set 2. 
(Best viewed in color)} 
\label{figut12}  
\vspace{0mm}
  \end{figure*}

{\renewcommand{\arraystretch}{1.5}
\setlength{\tabcolsep}{8pt}
\begin{table*}[h]
\begin{center}
\caption{Performance comparison of the proposed method with  other baselines on UT-Interaction dataset (Set 1). } 
\label{ut11}
\begin{tabular}{ccccccccccc}
\hline
 
\multirow{2}{*}{Methods} &
      \multicolumn{10}{c}{Observation Ratio} \\
        &0.1&0.2&0.3 & 0.4 & 0.5 & 0.6 & 0.7 & 0.8 & 0.9 &1.0 \\
\hline
Sp\_Tp\_Rank&53.33\%&56.67\%&63.33\%&70.00\%&\textbf{85.00\%}&85.00\%&90.00\%&91.67\%&91.67\%&91.67\%\\
     Sp\_Tp\_St\_Avg&41.67\%&45.00\%&50.00\%&65.00\%&75.00\%&78.33\%&85.00\%&88.33\%&85.00\%&85.00\%\\
  Proposed Method &\textbf{55.00\%}&\textbf{60.00\%}&\textbf{66.67\%}&\textbf{78.33\%}&83.33\%&\textbf{86.67\%}&\textbf{93.33\%}&\textbf{93.33\%}&\textbf{95.00\%}&\textbf{93.33\%}\\

\hline
\end{tabular}
\end{center}
\end{table*}
}

{\renewcommand{\arraystretch}{1.5}
\setlength{\tabcolsep}{8pt}
\begin{table*}[h]
\begin{center}
\caption{Performance comparison of the proposed method with  other baselines on UT-Interaction dataset (Set 2). } 
\label{ut21}
\begin{tabular}{ccccccccccc}
\hline
 
\multirow{2}{*}{Methods} &
      \multicolumn{10}{c}{Observation Ratio} \\
        &0.1&0.2&0.3 & 0.4 & 0.5 & 0.6 & 0.7 & 0.8 & 0.9 &1.0 \\
\hline

 Sp\_Tp\_Rank&38.33\%&40.00\%&50.00\%&70.00\%&81.67\%&86.67\%&88.33\%&86.67\%&88.33\%&88.33\%\\
    Sp\_Tp\_St\_Avg&38.33\%&40.00\%&48.33\%&60.00\%&78.33\%&83.33\%&85.00\%&85.00\%&86.67\%&83.33\%\\
  Proposed Method &\textbf{46.67\%}&\textbf{48.33\%}&\textbf{55.00\%}&\textbf{71.67\%}&\textbf{83.33\%}&\textbf{91.67\%}&\textbf{91.67\%}&\textbf{90.00\%}&\textbf{91.67\%}&\textbf{91.67\%}\\

\hline
\end{tabular}
\end{center}
\end{table*}
}

 {\renewcommand{\arraystretch}{1.5}
\setlength{\tabcolsep}{7.5pt}
\begin{table*}[h]
\begin{center}
\caption{Performance comparison of the proposed method with individual models on UT-Interaction dataset  (Set 1). } 
\label{ut12}
\begin{tabular}{ccccccccccc}
\hline
 
\multirow{2}{*}{Methods} &
      \multicolumn{10}{c}{Observation Ratio} \\
        &0.1&0.2&0.3 & 0.4 & 0.5 & 0.6 & 0.7 & 0.8 & 0.9 &1.0 \\
\hline

    Spatial&30.00\%&30.00\%&31.67\%&45.00\%&50.00\%&50.00\%&61.67\%&70.00\%&68.33\%&65.00\%\\
    Temporal \cite{ke2016human}&45.00\%&53.33\%&61.67\%&71.67\%&81.67\%&\textbf{86.67\%}&86.67\%&88.33\%&90.00\%&90.00\%\\
    Spatial Structural &33.33\%&35.00\%&36.67\%&48.33\%&56.67\%&65.00\%&66.67\%&70.00\%&73.33\%&73.33\%\\
    Temporal Structural&30.00\%&36.67\%&55.00\%&70.00\%&76.67\%&85.00\%&86.67\%&88.33\%&86.67\%&85.00\%\\
     Proposed Method &\textbf{55.00\%}&\textbf{60.00\%}&\textbf{66.67\%}&\textbf{78.33\%}&\textbf{83.33\%}&\textbf{86.67\%}&\textbf{93.33\%}&\textbf{93.33\%}&\textbf{95.00\%}&\textbf{93.33\%}\\

\hline
\end{tabular}
\end{center}
\end{table*}
}

 {\renewcommand{\arraystretch}{1.5}
\setlength{\tabcolsep}{7.5pt}
\begin{table*}[h]
\begin{center}
\caption{Performance comparison of the proposed method with individual models on UT-Interaction dataset (Set 2). } 
\label{ut22}
\begin{tabular}{ccccccccccc}
\hline
 
\multirow{2}{*}{Methods} &
      \multicolumn{10}{c}{Observation Ratio} \\
        &0.1&0.2&0.3 & 0.4 & 0.5 & 0.6 & 0.7 & 0.8 & 0.9 &1.0 \\
\hline
  Spatial&38.33\%&38.33\%&41.67\%&55.00\%&58.33\%&65.00\%&66.67\%&70.00\%&73.33\%&75.00\%\\
     Temporal \cite{ke2016human} &33.33\%&33.33\%&55.00\%&68.33\%&\textbf{85.00\%}&88.33\%&88.33\%&\textbf{90.00\%}&90.00\%&90.00\%\\
    Spatial Structural &38.33\%&38.33\%&41.67\%&46.67\%&46.67\%&55.00\%&60.00\%&56.67\%&56.67\%&58.33\%\\
    Temporal Structural& 36.67\%&41.67\%&\textbf{61.67\%}&\textbf{71.67\%}&78.33\%&80.00\%&78.33\%&78.33\%&80.00\%&81.67\%\\
   Proposed Method &\textbf{46.67\%}&\textbf{48.33\%}& 55.00\% &\textbf{71.67\%}&83.33\%&\textbf{91.67\%}&\textbf{91.67\%}&\textbf{90.00\%}&\textbf{91.67\%}&\textbf{91.67\%}\\

\hline
\end{tabular}
\end{center}
\end{table*}
}

\subsubsection{Results on the UT-Interaction Dataset}
 
There is no training/testing split with this dataset. The performance is measured using leave-one-sequence-out cross validation, i.e., for each set,  54 videos performed by 9 groups of actors are used for training and the remaining  sequences of 1 group of actors are used for cross validation. The model is validated  10 times and the averaged results are reported as the model performance.

    The proposed method  is compared with other methods, including IBoW \cite{Ryoo2011} DBoW \cite{Ryoo2011},  HM  \cite{lan2014hierarchical},   SC \cite{cao2013recognize} MSSC \cite{cao2013recognize},   and   MMAPM \cite{kong2015max}, and the results are shown in Figure \ref{figut12}.    
 It can be seen that the  proposed method  achieves superior results over other methods  in 7 out of 10 observation ratios for both Set 1 and Set 2. When the observation ratio is 0.1, the accuracy of the proposed method on Set 1 is about 55.0\%. The improvement compared with the best previous result (\ie, MMAPM \cite{kong2015max}) is about 8.33\%.   The improvements of the proposed method on Set 2 are more significant after an observation ratio of  0.4.  When testing with  half the length  of the videos (\ie,   observation ratio of 0.5), the proposed  method achieves an impressive 83.3\%   accuracy.  Compared to  MMAPM (75.0\%) \cite{kong2015max},  the improvement is 8.3\%. 
  
  Table \ref{ut11} and Table \ref{ut21} shows the comparison of the proposed method with  the baseline methods Sp\_Tp\_Rank and Sp\_Tp\_St\_Avg on Set 1 and Set 2. It can be seen that the proposed method outperforms the   baseline methods in 9 out of 10 observation ratios in Set 1 and all cases in Set 2 of the UT-Interaction dataset.   
  The comparison of the proposed method   with individual models are shown in  Table \ref{ut12} and Table \ref{ut22}. The weights of  the spatial, temporal, spatial structural and temporal structural models  learned by the ranking score fusion method are  $[0.16 ,  0.31 ,  0.23,    0.30]$  
  on Set 1, and   $[ 0.22, 0.25,   0.18, 0.35]$ on Set 2. The fusion results outperform  the performances of individual models in all cases in Set 1 and 8 out of 10 cases in Set 2. This clearly demonstrates the superiority of the proposed structural models and ranking method.
   

\section{Conclusion}

In this paper, we proposed  novel structural models  to uncover the contextual dependencies and salient information   for interaction prediction. The structural models are learned by using LSTM networks to process the sequence of global and local contexts.  
We also proposed a novel ranking score fusion method to determine the optimal weights of the spatial, temporal and   structural models,    to  effectively combine their complementary strengths. 
The proposed method was compared with previous works on two interaction datasets and has achieved superior performance.
We also performed an ablative analysis and compared the proposed method with a baseline that does not include
the structural models and a baseline that does not use the proposed ranking score fusion method. 
Experimental results   show
the benefits of the proposed framework, particularly the structural models and the fusion method.

\section*{Acknowledgment}
This work was partially supported by Australian Research Council grants DP150100294, DP110103336, and DE120102960.

\ifCLASSOPTIONcaptionsoff
  \newpage
\fi



%
\bibliographystyle{IEEEtran}
\bibliography{p1ac}

%


\begin{IEEEbiographynophoto}{Qiuhong Ke}
Qiuhong Ke received her M.S. degree from Beijing Forestry University, China in the area of computer vision.  She is currently working on her Ph.D. degree in Computer Science at the University of Western Australia. Her research interests include action recognition and deep learning.
\end{IEEEbiographynophoto}

\begin{IEEEbiographynophoto}{Mohammed Bennamoun}
Mohammed Bennamoun received the M.S. degree in control theory from Queens University, Kingston, Canada, and the Ph.D. degree in computer vision from Queens University/Queensland University of Technology, Brisbane, Australia. He is currently a Winthrop Professor with The University of Western Australia, Australia. He has authored over 300 journal and conference publications. His research interests include control theory, robotics, object recognition, artificial neural networks, signal/image processing, and computer vision. He was selected to give conference tutorials at the European Conference on Computer Vision (ECCV) and the International Conference on Acoustics Speech and Signal Processing. He organized several special sessions for conferences, e.g., the IEEE International Conference in Image Processing. He also contributed in the organization of many local and international conferences. He served as a Guest Editor of a couple of special issues in International journals, such as the International Journal of Pattern Recognition and Artificial Intelligence.
\end{IEEEbiographynophoto}

\begin{IEEEbiographynophoto}{Senjian An}
Senjian An received his B.S degree from Shandong University, the M.S. degree from the Chinese Academy of Sciences, and the Ph.D. degree from Peking University, China. He is currently a Research Fellow at the School of Computer Science and Software Engineering, The University of Western Australia. His research interests include machine learning, image processing, object detection and recognition.
\end{IEEEbiographynophoto}

\begin{IEEEbiographynophoto}{Ferdous   Sohel}
Ferdous  Sohel received the Ph.D. degree from Monash University, Australia, in 2009. He is currently a Senior Lecturer of Information Technology with Murdoch University, Australia. He was a Research Assistant Professor/Research Fellow with the School of Computer Science and Software Engineering, The University of Western Australia from 2008 to 2015.   His research interests include computer vision, image processing, pattern recognition, multimodal biometrics, scene understanding, robotics, and video coding. He is a member of Australian Computer Society. He is a recipient of prestigious Discovery Early Career Research Award by the Australian Research Council. He is also a recipient of the Early Career Investigators Award and the Best Ph.D. Thesis Medal from Monash University.
\end{IEEEbiographynophoto}

\begin{IEEEbiographynophoto}{Farid Boussaid}
Farid Boussaid received the M.S. and Ph.D. degrees in microelectronics from the National Institute of Applied Science (INSA), Toulouse, France, in 1996 and 1999 respectively. He joined Edith Cowan University, Perth, Australia, as a Postdoctoral Research Fellow, and a Member of the Visual Information Processing Research Group in 2000. He joined the University of Western Australia, Crawley, Australia, in 2005, where he is currently a Professor. His current research interests include smart CMOS vision sensors and image processing.
\end{IEEEbiographynophoto}




\end{document}